\documentclass[10pt,a4paper]{article}
\usepackage[T1]{fontenc}
\usepackage[utf8]{inputenc}
\usepackage{lmodern,textcomp}
\usepackage[british]{babel}
\usepackage[a4paper,margin=20mm]{geometry}
\usepackage{amsmath,amssymb}
\usepackage{graphicx}
\usepackage{booktabs,longtable,array,calc}
\usepackage{pdflscape}
\usepackage{caption}
\usepackage[section]{placeins}
\usepackage{microtype}
\usepackage[super,sort&compress]{natbib}
\usepackage{xurl}
\usepackage[hidelinks,unicode]{hyperref}

\DeclareUnicodeCharacter{2265}{\ensuremath{\geq}}
\DeclareUnicodeCharacter{2264}{\ensuremath{\leq}}
\DeclareUnicodeCharacter{2212}{\ensuremath{-}}
\DeclareUnicodeCharacter{0394}{\ensuremath{\Delta}}
\DeclareUnicodeCharacter{03C1}{\ensuremath{\rho}}
\DeclareUnicodeCharacter{03B1}{\ensuremath{\alpha}}
\DeclareUnicodeCharacter{207B}{\textsuperscript{-}}
\DeclareUnicodeCharacter{2070}{\textsuperscript{0}}

\renewcommand{\arraystretch}{1.15}

\makeatletter
\newcommand{\rsccontrol}{\immediate\write\@auxout{\string\citation{rsc-control}}}
\makeatother

\providecommand{\supplementtitle}{}
\title{\supplementtitle From Benchmarks to the Bench:\\
Preference, Correspondence and Selectivity in Bidirectional Drug–Target Screening}
\author{Jiangjie Qiu\textsuperscript{1,\textdagger}, Chengrui Xu\textsuperscript{1,\textdagger}, Xiaonan Wang\textsuperscript{1,*}\\[0.8em]
\parbox{0.94\textwidth}{\centering\small
\textsuperscript{1}Beijing Key Laboratory of Artificial Intelligence for Advanced Chemical Engineering Materials,
State Key Laboratory of Chemical Engineering and Low-Carbon Technology,
Department of Chemical Engineering, Tsinghua University, Beijing 100084, China.\\[0.4em]
\textsuperscript{\textdagger}These authors contributed equally to this work and share first authorship.\\[0.4em]
\textsuperscript{*}Corresponding author: Xiaonan Wang,
email: \href{mailto:wangxiaonan@tsinghua.edu.cn}{wangxiaonan@tsinghua.edu.cn}.}}
\date{}

\begin{document}
\maketitle
\begin{abstract}
Identifying active molecules for known targets and potential targets for existing drugs is central to drug discovery and repurposing. Drug–target interaction (DTI) models can rank vast numbers of drug–target pairs, but limited experimental capacity restricts testing to a small number of candidates. We compared the bidirectional recommendations of 13 released systems across 711 molecular structures of FDA-approved drugs and 614 human proteins. For the same target or drug, pairs of models produced completely disjoint Top-10 molecule and protein lists in 77.8\% and 70.1\% of comparisons, respectively. Even where recommendations overlapped, agreement between some systems largely reflected recurring preferences for the same candidates. Published inhibition measurements for 79 molecules and 65 kinases revealed that some systems enriched active candidates, yet their hit yields changed little when recommendations were matched to different drugs or targets, whereas other systems lost substantially more hits. Some of the former systems also rarely ranked both molecules and kinases correctly when two drugs favoured different kinases. We further retrained the same architecture on the same data, with the training objective fixed and random seeds varied. Despite similar test performance, mean pairwise Top-10 overlap on a separate screening catalogue was about two candidates. Our study moves DTI model evaluation beyond benchmark performance to assess whether model predictions can guide meaningful experimental decisions in drug discovery and target identification.
\end{abstract}

\section{Introduction}\label{introduction}

Mapping interactions between small molecules and biological macromolecules underpins chemogenomics and drug discovery. Drug–target interaction (DTI) prediction is used in forward ligand screening to identify molecules active against a given protein target \citep{lyu2019ultra,sadybekov2023computational}. Reverse target screening asks which proteins a known active molecule or approved drug may act on. Identifying these targets can suggest new uses for a drug or help explain off-target effects \citep{pushpakom2019drug,schenone2013target}. In molecule selection, the protein is the query and molecules are the candidates. In protein selection, these roles are reversed. DTI models can evaluate vast numbers of possible molecule–protein pairs \citep{conplex,drugclip_paper,graphban_paper,chemltf_paper,evidti_paper,probematch_paper,he2025}, but experimental budgets and throughput limit how many can be tested, even in emerging self-driving laboratories \citep{burger2020robotic,szymanski2023alab,szymanski2023correction}. The usefulness of a candidate list depends on the experimental objective. For an initial screen, measurable activity may be the main criterion. Target identification asks which proteins a molecule acts on, while selective probe development also requires distinguishing its activity on the intended target from activity on alternatives. The pairs selected for testing determine which new measurements enter the next round of a closed-loop experiment \citep{burger2020robotic,szymanski2023alab}.

DTI models are commonly evaluated using aggregate prediction metrics or their ability to recover known interactions \citep{he2025,bagherian2021machine}. Previous studies have shown that predictions can be influenced by ligand and target preferences, as well as biases in training and evaluation datasets \citep{wallach2018most,chen2019hidden,tapb,glitter}. For example, two systems may recommend the same molecule because both rank it highly for many targets. We therefore reassign one system's recommendation lists across queries to measure how much agreement remains. Finding active candidates does not establish whether a model's recommendations yield more hits for their intended drug or target than they would if assigned to other queries. Understanding these distinctions requires examining both the candidates selected for testing and their measured outcomes, rather than relying on aggregate prediction performance alone.

Selectivity requires comparing activity across alternative partners. Kinase profiling studies have shown that compounds can differ substantially in their activities across targets \citep{davis2011comprehensive,klaeger2017target}. When two molecules and two proteins exhibit crossed measured preferences, a strict evaluation requires all four comparisons to be correct: the model must identify the preferred molecule for each protein and the preferred protein for each molecule. Recommendations may also change when a model is retrained. Studies of underspecification and predictive multiplicity have shown that models with similar aggregate performance can disagree on individual predictions \citep{damour_underspecification,marx_multiplicity,black2022model}. Whether such differences change the candidates selected for testing, and whether different selections lead to different measured outcomes, cannot be determined from aggregate test performance alone.

We address these questions through three complementary analyses (Fig. 1). Comparing 13 released systems on a common catalogue of 711 molecular structures of FDA-approved drugs and 614 human proteins, we find that most cross-system Top-10 comparisons share no candidates, while much of the agreement between some systems persists when complete recommendation lists are reassigned across queries. Using published single-concentration inhibition measurements for all \(79 \times 65\) molecule–kinase pairs in a subset of the Kinase Inhibitor Repurposing Hub (KIRHub) \citep{kirhub2026}, we find that some systems enrich active candidates yet recover nearly the same number of hits whether their recommendations are evaluated for their original queries or reassigned among queries. Other systems lose substantially more hits under the same comparison. Some systems showing the former pattern also rarely recover crossed preferences between molecules and kinases in both screening directions, revealing a marked difference between finding activity and correctly ranking selective pairings. We additionally examine how alternative candidate-selection rules affect these outcomes. To assess reproducibility, we repeated training with different random seeds while holding the data split, encoders, architecture and objective fixed. The models had similar test performance, but their Top-10 lists shared only about two candidates on average in pairwise comparisons on a separate screening catalogue. On the measured overlap between this catalogue and the kinase panel, no seed shows a clear advantage in mean hit yield.

A model may identify active candidates without reliably distinguishing the drug–target preferences relevant to a particular investigation, while similar predictive performance can lead to substantially different experimental choices. We evaluate these differences through activity yield, query-associated gain, recovery of crossed preferences and stability across training runs. These measures relate a model's recommendations to the biological comparisons required by the experiment.

\begin{figure}[p]
\centering
\includegraphics[width=\linewidth,height=0.62\textheight,keepaspectratio]{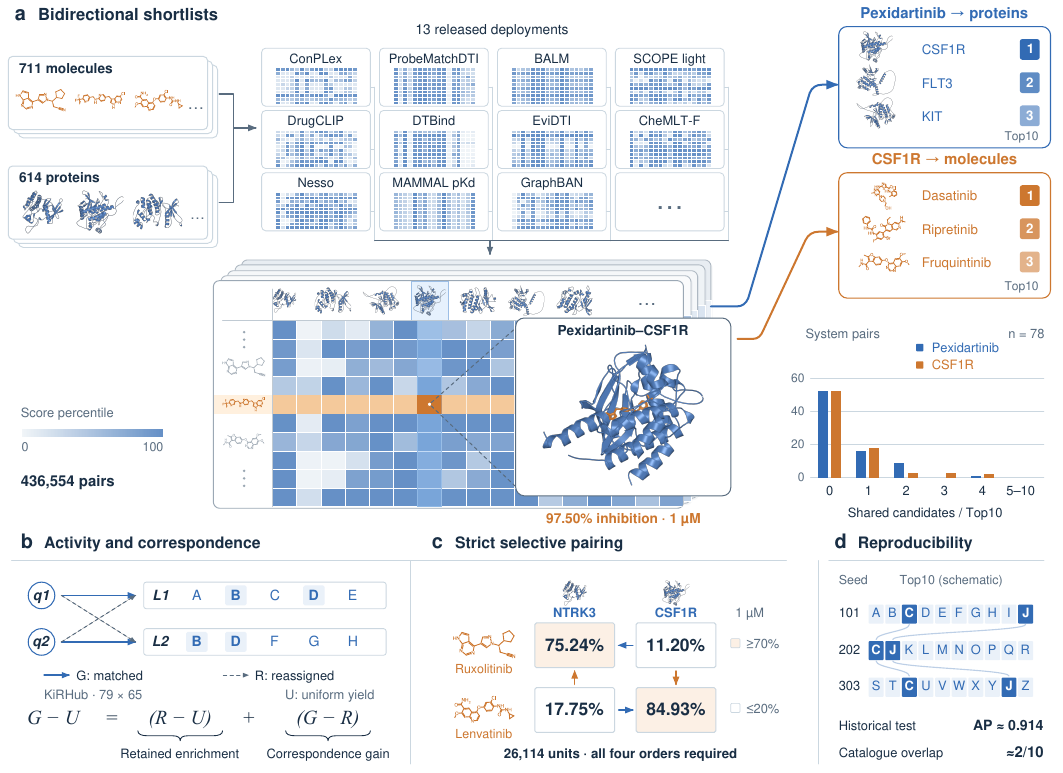}
\caption{\textbf{Shared drug--target candidates give rise to different experimental priorities.} \textbf{a,} Thirteen released prediction systems score the same 711 molecules and 614 proteins, comprising 436,554 pairs. Eleven named cards show the same 10-molecule × 17-protein excerpt. The ellipsis represents the remaining systems. Colours encode within-system score percentiles over all common pairs. Histograms show shared candidates across all 78 system pairs for pexidartinib (blue) and CSF1R (orange). No pair shares more than four candidates, regardless of boundary-tie resolution. The inset shows the engineered pexidartinib--CSF1R crystal complex (PDB 4R7H), rendered in PyMOL. The 97.50\% inhibition at 1 µM is a separate KIRHub measurement. Other protein cartoons use existing AlphaFold kinase-domain models. Chemical drawings use catalogue SMILES. \textbf{b,} Whole-list query reassignment preserves candidate order while changing query correspondence. Solid and dashed arrows show original and reassigned lists. Shading marks shared candidates. On the functional panel, excess Top-5 hit yield, \(G-U\), is decomposed into retained enrichment, \(R-U\), and correspondence gain, \(G-R\). Here, \(R\) averages within-fold reassignments and \(U\) is the uniform-selection reference. \textbf{c,} Ruxolitinib and lenvatinib against NTRK3 and CSF1R illustrate a crossed-selectivity unit. Preferred pairs reach ≥70\% inhibition and crossed pairs remain ≤20\% at 1 µM. Arrows indicate the four required strict orders. Ties fail. The 26,114 qualifying units share entities and are not independent observations. \textbf{d,} Three training seeds share the same data, encoders, architecture and objective. Mean labelled-test average precision (AP) is 0.9137. Mean pairwise Top-10 overlap on the separate 468 × 469 catalogue is 2.215 molecules or 1.885 proteins. The schematic lists share two candidates per seed pair. Structural reference: \href{https://www.rcsb.org/structure/4R7H}{PDB 4R7H} and \href{https://doi.org/10.1056/NEJMoa1411366}{primary publication}.}
\label{fig:1}
\end{figure}

\section{Results}\label{results}

\subsection{Bidirectional recommendations from released prediction systems}\label{bidirectional-recommendations-from-released-prediction-systems}

Under a common candidate pool and budget, released prediction systems proposed different experimental candidates for the same query. We compared thirteen systems using their supplied weights and specified input and output configurations (Table~\ref{tab:1}). The common matrix contained 711 molecules, 614 proteins and 436,554 pairs, with valid scores from every system. For each protein query, we selected the ten highest-ranked molecules. For each molecule query, we selected the ten highest-ranked proteins. Candidates were ranked using each system's own scores.

\begingroup
\renewcommand{\thetable}{1}
\setlength{\tabcolsep}{4pt}
\setlength{\LTcapwidth}{\linewidth}
\renewcommand{\arraystretch}{1.18}
\small
\begin{longtable}[]{@{}
  >{\raggedright\arraybackslash}p{(\linewidth - 2\tabcolsep) * \real{0.2200}}
  >{\raggedright\arraybackslash}p{(\linewidth - 2\tabcolsep) * \real{0.7800}}@{}}
\caption{\textbf{Released prediction systems and evaluated outputs.} Each row represents one system under its specified inference configuration. Ensembles of released models are treated as single systems. Scores differ in training objective and scale and are used to rank candidates within each system.}\label{tab:1}\tabularnewline
\toprule\noalign{}
\begin{minipage}[b]{\linewidth}\raggedright
System and source
\end{minipage} & \begin{minipage}[b]{\linewidth}\raggedright
Evaluated configuration or output
\end{minipage} \\
\midrule\noalign{}
\endfirsthead
\toprule\noalign{}
\begin{minipage}[b]{\linewidth}\raggedright
System and source
\end{minipage} & \begin{minipage}[b]{\linewidth}\raggedright
Evaluated configuration or output
\end{minipage} \\
\midrule\noalign{}
\endhead
\bottomrule\noalign{}
\endlastfoot
ConPLex \citep{conplex} & Contrastive protein--molecule embeddings. BindingDB-trained model \\
DrugCLIP \citep{drugclip_paper} & Pocket--molecule contrastive retrieval. Mean raw cosine similarity across six released fold models, using one pocket per protein \\
Nesso-1 \citep{nesso_report} & Co-folding model. Binary binding-classification score with five recycling steps \\
ProbeMatchDTI \citep{probematch_paper} & Multi-scale biochemical pattern matching. Pairwise classification score \\
DTBind \citep{dtbind_paper} & Molecular-recognition model. Binding-occurrence score using reconstructed surface features \\
MAMMAL pKd \citep{mammal_paper} & Multimodal biomedical language model. BindingDB pKd output: the negative base-10 logarithm of the dissociation constant in molar units \\
BALM \citep{balm_paper} & Protein and ligand language models. BindingDB affinity model \\
EviDTI \citep{evidti_paper} & Evidential deep learning. DrugBank classification model \\
GraphBAN \citep{graphban_paper} & Graph-based interaction prediction. Mean positive-class softmax probability across 20 released case-study models \\
SCOPE light \citep{scope} & Protein--molecule feature fusion. Mean sigmoid probability across five models in the released ``Total'' set \\
CheMLT-F \citep{chemltf_paper} & Multitask transformer fusion. KIBA prediction output \\
ADME-DTI \citep{admedti_paper} & Multi-descriptor ensemble. Released ``BindingDB Combined'' classifier \\
UniDTI \citep{unidti_paper} & Multimodal, multi-scale model. Sigmoid-transformed binary classification logit \\
\end{longtable}

\endgroup

We compared all 78 distinct pairs formed by the 13 systems. Across all queries, these pairs yielded 47,892 molecule-selection and 55,458 protein-selection comparisons. Top-10 lists were completely disjoint in 77.8\% of molecule-selection comparisons and 70.1\% of protein-selection comparisons. Thus, in most comparisons, choosing a different system changed all ten candidates proposed for the same query and testing budget. For each system pair, we also averaged the number of shared Top-10 candidates across queries. The medians of these 78 averages were 0.2289 molecules and 0.3390 proteins, both below one shared candidate per ten recommendations.

To assess sensitivity to specific input and output configurations, we repeated the analysis after excluding DrugCLIP, DTBind, GraphBAN and CheMLT-F, while keeping the candidate pools unchanged. In this nine-system subset, Top-10 lists were completely disjoint in 78.49\% of molecule-selection comparisons and 67.39\% of protein-selection comparisons (Supplementary S3). We next reassigned one system's complete recommendation lists across queries, preserving the candidates and their order within each list. Comparing the observed overlap with the exact reassignment reference quantifies how much agreement is retained through recurring candidate preferences and how much is associated with the original query correspondence.

\subsection{Persistent candidate preferences and query-dependent agreement}\label{persistent-candidate-preferences-and-query-dependent-agreement}

\begin{figure}[p]
\centering
\includegraphics[width=\linewidth,height=0.62\textheight,keepaspectratio]{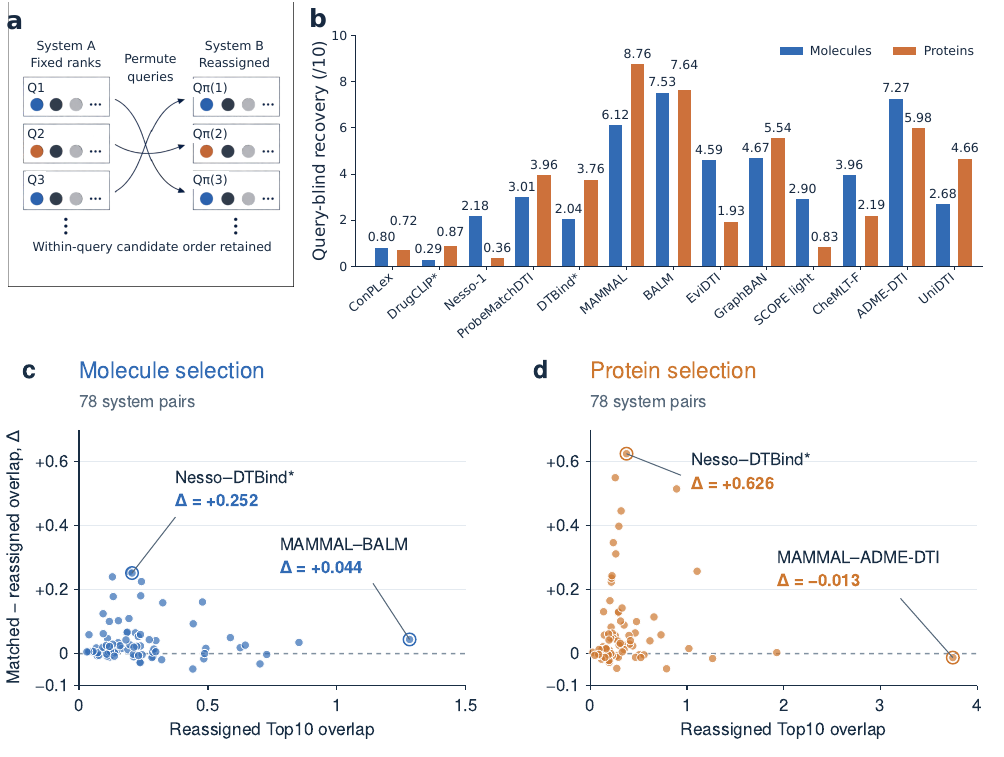}
\caption{\textbf{Persistent candidate preference and query-dependent agreement.} \textbf{a,} Query correspondence is reassigned while the candidate ordering within each list is retained. The exact reassignment reference averages all query permutations, including fixed points. \textbf{b,} Mean Top-10 recovery by candidate preferences fitted on other query groups, shown for thirteen released systems in molecule selection (blue) and protein selection (orange). Values are shared candidates per ten recommendations. \textbf{c,d,} Reassigned Top-10 overlap, \(R\), versus the query-aligned agreement increment, \(\Delta\), for all 78 unordered system pairs when selecting molecules (c) or proteins (d). The horizontal dashed line marks zero increment. Both panels use the same vertical scale. Horizontal ranges differ. Each point summarises a system pair over the common query pool. System pairs share models and are not independent observations. Asterisks identify the DrugCLIP and DTBind input configurations described in Supplementary S1.}
\label{fig:2}
\end{figure}

We first measured how much of each system's recommendation could be recovered from persistent candidate preferences. In molecule selection, BALM's ten most frequent molecules occupied 84.9\% of its Top-10 slots, compared with 7.8\% for DrugCLIP. In protein selection, the ten most frequent proteins from MAMMAL pKd occupied 87.6\%. We fitted a fixed query-blind ranking from mean candidate percentile ranks in other query groups, without using the current query. On held-out queries, this ranking recovered 8.76 of the Top-10 proteins from MAMMAL pKd and 7.53 of the Top-10 molecules from BALM. The corresponding recoveries for Nesso-1 were 0.36 proteins and 2.18 molecules. Persistent preferences thus reproduced different fractions of the systems' recommendations (Fig.~\ref{fig:2}b and Supplementary Fig. S1).

To separate sources of agreement, we preserved candidate identity and each list's internal order while reassigning one system's query correspondence. For MAMMAL pKd--BALM, the molecule-selection correlation was 0.4217, with an exact reassignment reference of 0.4201. The corresponding protein-selection values were 0.3789 and 0.3784. MAMMAL pKd--ADME-DTI shared 3.736 Top-10 proteins on average, versus a reference of 3.748. Most agreement for these pairs remained after query reassignment (Fig.~\ref{fig:2}c,d).

Across all 78 pairs, median query-aligned agreement increments for Spearman correlation were 0.00233 and 0.00177 in the two directions, and median Top-10 increments were 0.0181 and 0.0252. Some pairs showed stronger query alignment. For Nesso-1--DTBind protein rankings, the whole-list correlation was \(-0.0974\). Their Top-10 lists shared 1.000 candidates on average, compared with a reference of 0.3745, an increment of 0.6255. MAMMAL pKd--BALM also showed similar observed and reference correlations on a separate 128 × 128 catalogue sharing no molecule or protein identities with the common catalogue (Supplementary S3 and Table S7). The increment quantifies query-aligned excess agreement between systems. It does not measure either system's query dependence alone, because systems with different query-dependent responses can also have a small increment. These label-free analyses distinguish sources of agreement, while measured activity is needed to assess whether query correspondence accompanies additional hits.

\subsection{Activity enrichment and correspondence gain}\label{activity-enrichment-and-correspondence-gain}

The retrospective functional panel separated activity enrichment from correspondence gain. The selected KIRHub subset comprised 5,135 single-concentration measurements for 79 molecules and 65 kinases \citep{kirhub2026}. A hit was defined as at least 70\% inhibition at 1 µM. In total, 628 pairs met this criterion. Hit yield was the mean number of hits among the selected candidates per query. With five candidates per query, uniform selection gave an expected \(U=0.611\) hits. All queries were retained, including 15 molecules and 10 proteins without measured hits.

We asked whether each list yielded more hits for its original query than when reassigned to other queries. Matched hit yield, \(G\), was the mean Top-5 hit yield under the original query correspondence. Reassigned hit yield, \(R\), was the exact mean after complete recommendation lists were reassigned within evaluation folds. The identity \(G-U=(R-U)+(G-R)\) separates excess yield into enrichment retained after reassignment, \(R-U\), and correspondence gain, \(G-R\). The latter measures the additional hit yield associated with the original query correspondence within this panel, budget and reassignment scheme.

MAMMAL pKd and BALM protein lists yielded 1.114 and 1.329 hits per five candidates, above the uniform reference, but their correspondence gains were only 0.025 and \(-0.001\). Nesso-1 yielded 2.354 hits with a gain of 1.466, and SCOPE light yielded 1.962 with a gain of 1.069. UniDTI yielded 0.908 molecule hits and 0.456 protein hits, the latter below the uniform expectation of 0.611. Its correspondence gains were 0.044 and 0.073, with both conditional resampling intervals including zero. Figure 3 and Supplementary Table S8a provide the complete comparison. Activity enrichment with low correspondence gain for MAMMAL pKd and BALM also persisted in the nine-system subset.

To assess dependence on pairs found in the two audited BindingDB source extracts, we evaluated a complete 57-molecule × 65-kinase subpanel after excluding their exact pair matches. MAMMAL pKd and BALM protein lists yielded 1.000 and 1.246 hits, above the uniform reference of 0.563, with correspondence gains of 0.030 for both. Nesso-1 and SCOPE light retained protein-selection gains of 1.429 and 0.936. In molecule selection, MAMMAL pKd yielded 0.415 hits, below the uniform reference, whereas BALM yielded 0.815 with a gain of −0.052. Nesso-1 and SCOPE light had molecule-selection gains of 1.258 and 0.217, respectively (Supplementary Tables S1d and S8b).

A query-blind baseline fitted to mean candidate preferences on development queries yielded 1.292 molecule hits and 1.291 protein hits on held-out queries (Supplementary Table S10). The ranking was fitted on development queries that were separate from the evaluation queries. Unlike \(R\), which uses evaluation measurements to interpret completed recommendations, this baseline generates recommendations from development-query scores alone. It therefore demonstrates activity enrichment without conditioning on the current query. Neither enrichment nor correspondence gain, however, tests whether scores recover reciprocal preferences among alternative partners.

\begin{figure}[p]
\centering
\includegraphics[width=\linewidth,height=0.62\textheight,keepaspectratio]{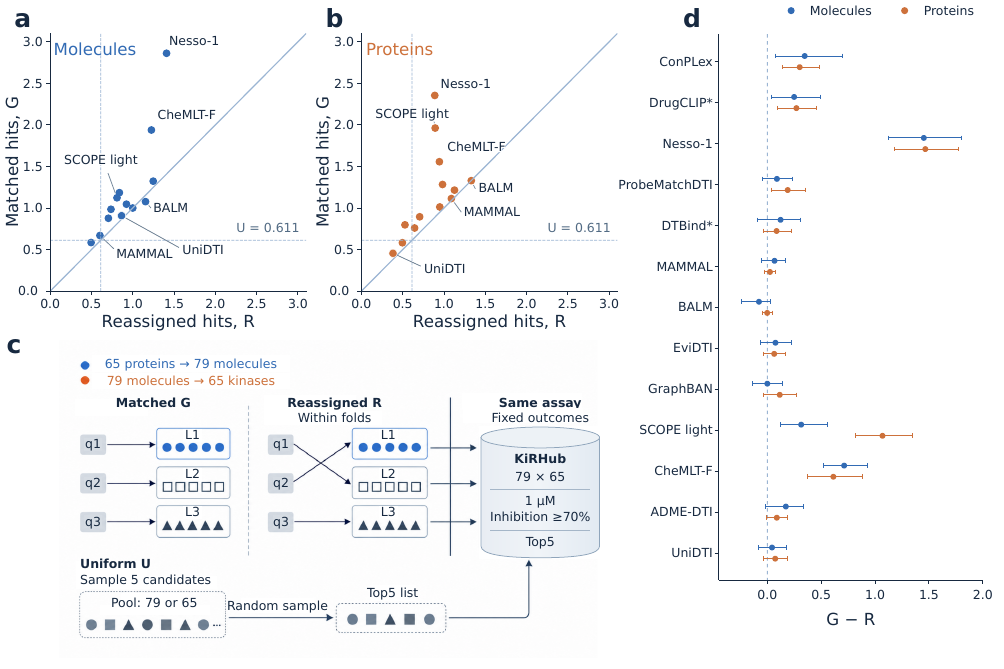}
\caption{\textbf{Activity enrichment and correspondence gain differ across systems.} \textbf{a,b,} Reassigned hit yield, \(R\), and matched hit yield, \(G\), for thirteen systems selecting molecules (a, blue) or proteins (b, orange). Hit yield is the number of hits per five candidates at ≥70\% inhibition at 1 µM on the complete 79 × 65 KIRHub panel. Dashed lines mark the uniform expectation, \(U=0.611\), and the diagonal marks \(G=R\). Labels identify selected systems, including UniDTI. \textbf{c,} Matched, reassigned and uniform references use the same measurements. Reassignment preserves each complete list and its internal candidate dependence. \textbf{d,} Correspondence gain, \(G-R\), in both directions. Error bars are conditional 95\% percentile intervals from 2,000 query-group resamples with fitted selection rules, reassignment references and candidate identities fixed. These intervals do not include uncertainty from measurement error, refitting prediction models or selection rules, or changes in the candidate pool. Fractional membership preserves the budget at tied boundaries. UniDTI scores were computed for the original 79 × 68 panel in fixed batches; the primary 79 × 65 panel retained those scores without recomputing the batches. This is a retrospective evaluation against an inhibition endpoint. Asterisks identify the DrugCLIP and DTBind input configurations described in Supplementary S1.}
\label{fig:3}
\end{figure}

\subsection{Strict crossed selectivity}\label{strict-crossed-selectivity}

We next tested strict crossed selectivity in measured two-molecule/two-protein units. A unit qualified when both preferred pairs had at least 70\% inhibition at 1 µM and both crossed pairs had at most 20\% inhibition. Strict recovery required all four molecule-wise and protein-wise orders to be correct, with ties counted as failures. The primary panel contained 26,114 qualifying units, involving 64 molecules and 55 proteins. Figure 4a illustrates the criterion with ruxolitinib, lenvatinib, NTRK3 and CSF1R.

\begin{figure}[p]
\centering
\includegraphics[width=\linewidth,height=0.62\textheight,keepaspectratio]{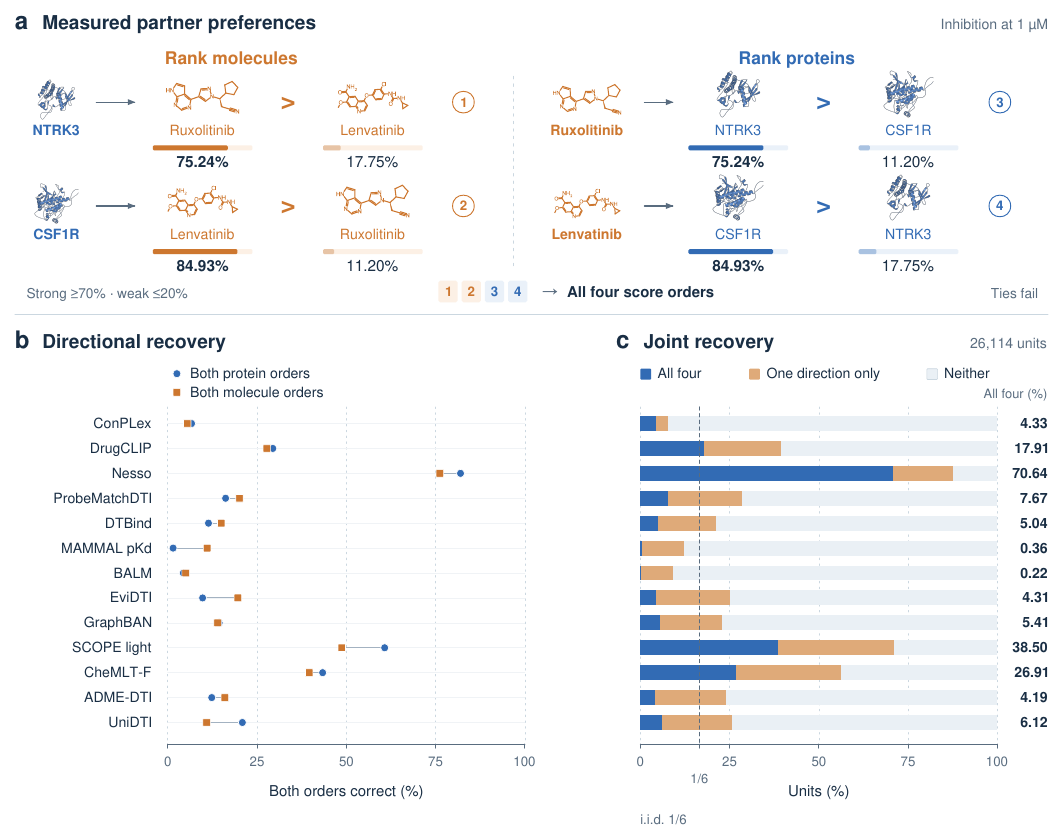}
\caption{\textbf{Strict crossed selectivity requires both directional preferences to reverse correctly.} \textbf{a,} Measured ruxolitinib/lenvatinib--NTRK3/CSF1R inhibition illustrates two molecule orders for fixed protein queries (orange) and two protein orders for fixed molecule queries (blue). Bars show inhibition on a common 0--100\% scale. Preferred pairs have ≥70\% inhibition and crossed pairs ≤20\% at 1 µM. Scores must recover all four strict orders; ties fail. Chemical drawings use source SMILES, and protein cartoons depict the corresponding kinase domains. \textbf{b,} Fractions of the 26,114 qualifying units with both protein orders correct (blue circles) or both molecule orders correct (orange squares). Symbols are centred on the same system row and connected horizontally. \textbf{c,} Units partitioned into all four orders correct (blue), exactly one successful direction (muted orange), or neither direction successful (pale grey). A successful direction requires both of its orders; ``neither'' does not imply that every individual order is wrong. The right column gives strict recovery rates. The dashed line marks the \(1/6\) reference for independent, identically distributed continuous scores. All systems use the same units involving 64 molecules and 55 proteins. Units share entities and are not independent. UniDTI retains scores from the original 79 × 68 inference batches (Methods).}
\label{fig:4}
\end{figure}

Across the thirteen systems, strict recovery ranged from 0.22\% to 70.64\%, with a median of 5.41\% (Fig.~\ref{fig:4}c and Supplementary Table S9). MAMMAL pKd and BALM recovered 0.36\% and 0.22\%, corresponding to 93 and 58 units. UniDTI, SCOPE light and Nesso-1 recovered 6.12\%, 38.50\% and 70.64\%, or 1,599, 10,053 and 18,448 units, respectively (Fig.~\ref{fig:4} and Supplementary Table S9). These rates describe ordering recovery among qualifying crossed-selectivity units. Although MAMMAL pKd and BALM enriched active candidates in protein selection, they rarely recovered reciprocal preferences within these units.

The reference recovery rates were zero for additive molecule-plus-protein scores and \(1/6\), approximately 16.7\%, for four independent, identically distributed continuous scores (Methods). These are mathematical references for score ordering, not significance thresholds. The measured units share molecules and proteins and are not independent observations.

\begin{figure}[p]
\centering
\includegraphics[width=\linewidth,height=0.62\textheight,keepaspectratio]{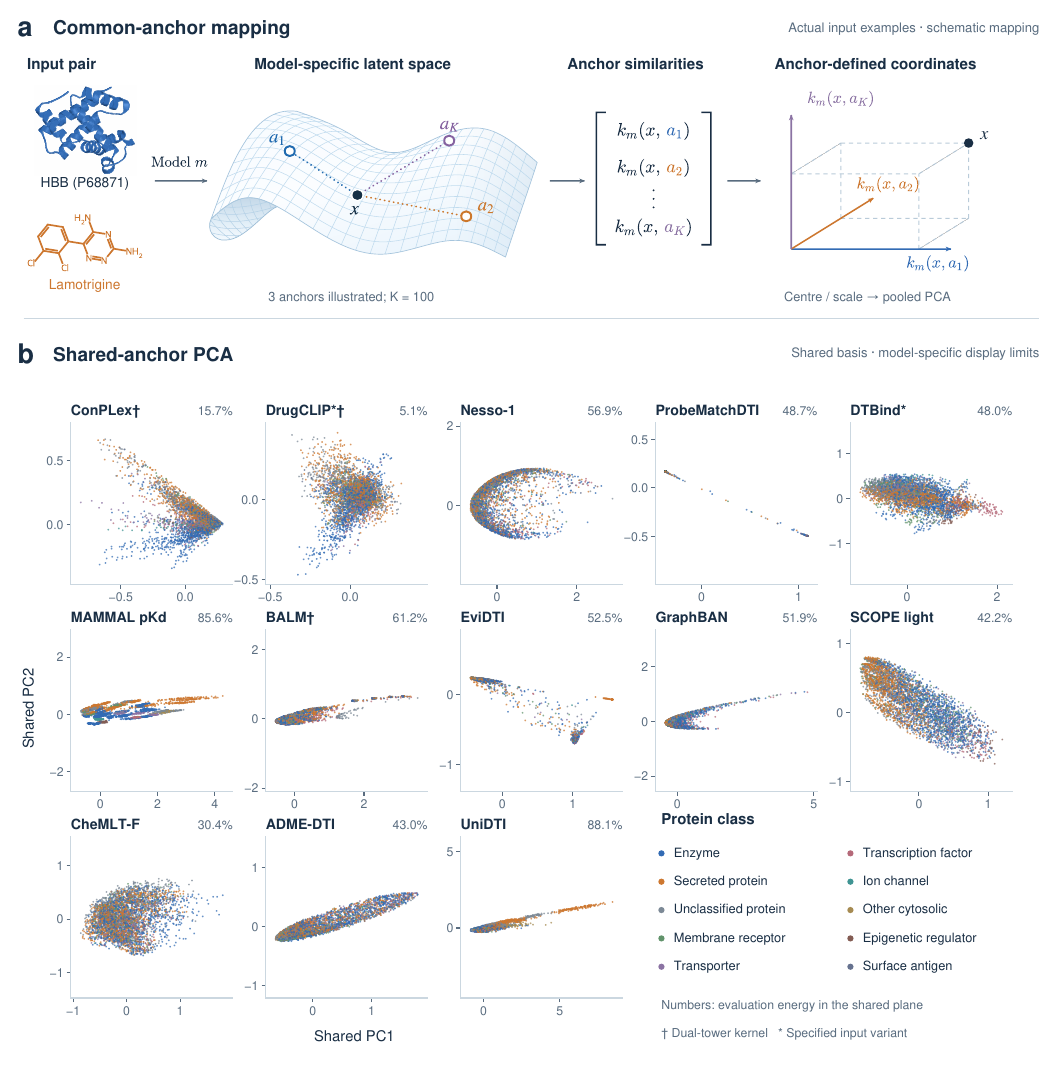}
\caption{\textbf{Common-anchor mapping reveals distinct representation geometries across prediction systems.} \textbf{a,} Schematic of the common-anchor mapping procedure illustrated with haemoglobin subunit beta (HBB, UniProt P68871) and lamotrigine. Each model \(m\) maps an input pair \(x\) into its model-specific latent space, where kernel similarities \(k_m(x, a_1), \dots, k_m(x, a_K)\) are evaluated against \(K = 100\) shared calibration anchor pairs (three anchors \(a_1, a_2, a_K\) illustrated). The resulting \(K\)-dimensional anchor-similarity vectors are centred and scaled using calibration statistics and projected onto a pooled principal component (PC) basis. \textbf{b,} Projection of the \(64 \times 64\) evaluation grid (4,096 pairs whose molecule and protein identities are separate from the calibration set) onto the first two shared principal components (Shared PC1 and Shared PC2) across all thirteen released systems. Points are coloured by target protein class across ten functional categories. Subplots use model-specific display limits to reveal internal structure across differing variance scales. Percentages above each subplot report the evaluation energy (the sum of squared anchor coordinates centred using the calibration mean) captured within the two-dimensional shared PC1--PC2 plane, ranging from 5.1\% for DrugCLIP to 88.1\% for UniDTI. Daggers (\(\dagger\)) denote dual-tower systems represented via product kernels over separate protein and ligand embeddings (ConPLex, DrugCLIP and BALM; Supplementary S6), and asterisks (\(*\)) indicate the specified input configurations for DrugCLIP and DTBind (Supplementary S1).}
\label{fig:5}
\end{figure}

To examine how representation differences relate to ranking differences, we compared model representations and the readouts that map them to scores (Fig.~\ref{fig:5} and Supplementary S6--S7). We used model-specific and common-anchor principal component analysis (PCA), where common-anchor mapping represented each pair by its similarities to 100 shared calibration pairs \citep{li2026platonic} (Fig.~\ref{fig:5}a). On a 64 × 64 evaluation grid with molecule and protein identities separate from calibration, systems occupied markedly different geometries in the shared anchor PC plane, with the two-dimensional plane capturing from 5.1\% (DrugCLIP) to 88.1\% (UniDTI) of anchor-space evaluation energy (Fig.~\ref{fig:5}b). Across system pairs, median centred kernel alignment (CKA), a measure of geometric similarity, was 0.02738 in the anchor space, and median local-neighbour overlap was 0.00417. Linear readouts fitted to calibration scores recovered median Top-10 overlaps of 5.95 molecules and 6.55 proteins from each system's own anchor coordinates, versus 1.84 and 1.95 from another system's coordinates.

We then tested how hit yield and strict recovery changed when pair-dependent information was altered within a system. On the functional panel, we scaled the doubly centred pair residual while keeping the downstream score-prediction layers fixed. This residual is the representation component remaining after removal of the overall mean and molecule and protein main effects. Residual removal could reduce strict recovery to zero while increasing the point estimates of Top-5 hit yield in both screening directions (Supplementary S7 and Table S13). These synthetic interventions illustrate that hit yield and strict crossed-selectivity recovery can respond differently to changes in pair-dependent information.

\subsection{Selection rules and hit yield}\label{selection-rules-and-hit-yield}

Because individual deployments produced disjoint recommendation lists shaped by persistent marginal preferences, we next tested how selection rules translate scores into a budget of five candidates per query. Standard consensus rules---such as rank averaging or voting---aim to reduce individual scoring errors \citep{Lacour2026DockM8,Agha2026MolDockLab}, but shared entity-level preferences may preserve marginal enrichment without improving reciprocal selectivity. Using five-fold grouped evaluation in each screening direction, we compared a development-selected single system, mean percentile ranks, Top-10 voting, and mean ranks adjusted for development-estimated candidate preferences. Out-of-fold hit yield was evaluated across all primary-panel queries, and strict recovery across the 832 fold-matched units (Methods and Supplementary S4).

Nesso-1 was selected in all five outer folds in both directions, yielding 2.862 molecule hits and 2.354 protein hits per query with 66.83\% strict recovery on these 832 units. Unadjusted consensus rules performed substantially worse: mean ranks yielded 1.969 and 1.728 hits (18.39\% strict recovery) and Top-10 voting yielded 1.951 and 1.785 hits (14.06\% strict recovery). Preference adjustment partially mitigated this drop, increasing yields to 2.138 and 1.987 hits, lifting strict recovery to 37.26\%, and raising correspondence gain over mean ranks from 0.551 to 0.798 for molecules and from 0.485 to 0.897 for proteins. Thus, while the development-selected single system outperformed the simple combinations tested here, discounting entity-level preferences substantially improved multi-system aggregation (Supplementary Table S10 and Fig. S4).

In a brief auxiliary check of within-system criteria, we also compared activity ranking with low-entropy selection, which favours predicted hit probabilities near zero or one. Hit-probability and inhibition-regression models were fitted from each deployment's native scalar score under simultaneous molecular-scaffold and protein-homology holdouts. At 25\% candidate coverage in molecule selection, low-entropy selection yielded lower selected-set inhibition MAE than activity ranking across all thirteen systems, yet reduced group-averaged positive retention below the 25\% uniform expectation in twelve systems (e.g., 2.31\% versus 78.75\% for Nesso-1; Supplementary S5 and Table S11). Lower prediction error on the selected set therefore did not imply greater activity recovery.

\subsection{Recommendation reproducibility across training seeds}\label{recommendation-reproducibility-across-training-seeds}

The cross-system comparisons involved different training sources and prediction settings. To isolate variation across repeated fits, we fixed the data split, feature encoders, architecture and objective and changed only the training seed. Seeds 101, 202 and 303 had mean labelled-test average precision (AP) of 0.9137 at step 5,000, ranging from 0.9076 to 0.9218. On a separate 468-molecule × 469-protein screening catalogue, seed pairs shared a mean of 2.215 molecules or 1.885 proteins per Top-10 list (Supplementary S8, Tables S28--S29 and Fig. S11).

Models selected by validation performance had mean AP of 0.9148 and Top-10 overlaps of 2.292 molecules and 2.071 proteins on the larger catalogue. Independent boundary-tie breaking within the same model gave a self-overlap reference of approximately 9.94. Both fixed-step and validation-selected comparisons therefore paired similar test performance with different proposed candidates for experimental testing. Additional controls varying ranking objectives, the measured pairs eligible for ranking supervision and sampling also changed recommendation agreement and query-level performance, without one objective dominating all evaluations (Supplementary S8 and Tables S14--S15).

To examine whether recommendation changes accompanied differences in hit yield, we evaluated the overlap between the screening catalogue and KIRHub. This panel contained 70 molecules, 65 kinases and 4,550 measurements. At five candidates per query, the mean absolute seed-pair difference per query was 0.323 molecule hits and 0.554 protein hits at step 5,000. All conditional 95\% resampling intervals for pairwise differences in mean hit yield included zero. The overlap panel contained 2,057 pairs used in training or validation-based model selection. After excluding them and reranking eligible candidates among the remaining 2,493 measurements, these intervals still included zero (Supplementary S9 and Tables S16 and S30). Recommendation overlap on the larger catalogue and mean hit yield on the overlap panel describe different outcomes in different evaluation sets. The latter did not identify a seed with a clear mean hit-yield advantage.

\section{Discussion}\label{discussion}

Under limited experimental budgets, moving drug--target interaction (DTI) evaluation from aggregate benchmarks to the experimental bench requires assessing which candidates a model prioritises and what their measured outcomes support. Existing comparisons focus on target recovery and aggregate screening performance \citep{he2025}, whereas studies of ligand bias, input permutation and target priors examine how single-entity properties shape predictions \citep{wallach2018most,glitter,tapb}. We connect these questions by examining how persistent candidate preference, query-specific correspondence, reciprocal selectivity and recommendation stability govern experimental prioritisation. Comparing released deployments on shared candidates and budgets evaluates recommendation behaviour under specified model configurations, while controlled retraining isolates variation within a fixed learning setup.

The value of persistent candidate preference depends on the experimental objective. Broadly active candidates can support initial activity discovery across several queries, whereas partner identification requires quantifying what the current query contributes beyond baseline preference. Query reassignment provides an operational reference: the query-aligned agreement increment, \(\Delta\), measures excess agreement between systems, whereas correspondence gain, \(G-R\), measures additional hits associated with the original query correspondence. Because correspondence gain alone does not establish selectivity against alternative partners, hit yield and correspondence gain should be reported separately, with selectivity evaluated through direct partner comparisons.

Strict crossed selectivity tests reciprocal partner preferences within measured crossed-selectivity units, although it does not estimate the prevalence of selective pairs in a candidate pool or replace concentration-dependent and broader target-profile measurements. To clarify why hit yield and partner ordering diverge across deployments, our common-anchor and readout analyses relate representation geometry to model scores and rankings, while synthetic residual scaling shows that removing pair-dependent information can reduce strict recovery to zero even when point estimates of Top-5 hit yield increase (Supplementary S6--S7 and Table S13). These diagnostics complement output-level evaluation without making representation similarity a substitute for measured recommendation performance.

Experimental objectives also govern how predictions should be aggregated or filtered by confidence. Multi-system aggregation should be judged by whether combining recommendations improves experimental outcomes: unadjusted rank averaging and voting preserved marginal enrichment without matching the best single system, whereas discounting development-estimated candidate preferences substantially improved both hit yield and strict recovery (Supplementary S4). Auxiliary-probability analyses likewise show that lower prediction error on a selected set does not imply greater activity discovery (Supplementary S5). Because low predictive entropy can reflect confidence in inactivity, exclusion-oriented filtering and hit discovery require distinct selection criteria matched to the experimental objective and budget.

Changes in recommended candidates across repeated training runs need not imply a loss of mean hit yield. List divergence within a fixed learning setup illustrates an experimental-prioritisation consequence of underspecification and predictive multiplicity \citep{damour_underspecification,marx_multiplicity,black2022model}. On the overlap panel, query-level outcomes varied across seeds, but conditional intervals for pairwise differences in mean hit yield all included zero (Supplementary S8--S9). These results support seed-dependent candidate selection without identifying a superior seed or establishing functional equivalence. Evaluation across repeated fits should therefore report both candidate identity and its measurable functional consequences, especially in closed-loop laboratories where divergent selections alter the observations available for subsequent updates \citep{burger2020robotic,szymanski2023alab}.

This study combines released deployments, a retrospective single-concentration kinase panel and one controlled learning setup into a decision-oriented evaluation framework. Its methodological contribution is to translate experimental objectives into distinct requirements: activity yield within budget, query-associated gain, conditional selectivity among alternative partners and recommendation stability across fits. Query reassignment, yield decomposition, crossed-order diagnostics and repeated training provide corresponding evaluation steps that can organise evaluation across other candidate pools and experimental endpoints, with the measures defined for their assay conditions. These diagnostics relate computational priorities to the experimental objective by assessing which candidates are selected and what partner relationships their measured outcomes can support.

\section{Methods}\label{methods}

\subsection{Candidate catalogue and prediction systems}\label{candidate-catalogue-and-prediction-systems}
The molecular catalogue comprised 720 unique structures representing 683 FDA-approved small-molecule drugs, including administered forms and ChEMBL-annotated active species. We retained drugs with defined structures and direct protein-action annotations, applied physicochemical eligibility criteria and deduplicated candidates by full InChIKey (Supplementary S1). Protein candidates came from 888 human single-protein targets with drug-mechanism annotations in ChEMBL 37. We excluded 143 G protein-coupled receptors, whose receptor-specific functional assays fell outside the screening scope, and 129 proteins longer than 1,000 residues to limit inference cost, retaining 616 proteins.

To compare recommendations over identical candidate pools, we required finite scores from all thirteen systems for every pair. Eight structures lacked ADME-DTI descriptors, and quinidine lacked a Nesso-1 prediction. DIO1 and TXNRD1 lacked usable DTBind surface inputs, and DIO1 also lacked a sequence-matched DrugCLIP pocket. Removing these entries from every system yielded a common matrix of 711 molecules × 614 proteins, comprising 436,554 pairs.

We evaluated released weights using the inputs and outputs specified in Table 1 and Supplementary Table S1a. Scores were oriented so that larger values ranked first, while retaining each method’s prediction endpoint and scale. DrugCLIP, GraphBAN, SCOPE light and ADME-DTI combined predictions from multiple released models, with each method’s ensemble evaluated as one system. DrugCLIP used raw cosine scores for one pocket per protein, DTBind used surface features reconstructed from sequence-matched AlphaFold structures, and GraphBAN used its released case-study ensemble. CheMLT-F used the KIBA regression output, a composite activity score whose rankings were evaluated against measured inhibition. A nine-system sensitivity analysis excluded these four configurations while retaining the candidate matrix. UniDTI used a fixed inference order and batch size because its scores depended on batch composition \citep{unidti_paper}. Deployment details, batch-composition sensitivity and source-data overlap checks are reported in Supplementary S1.

\subsection{Bidirectional ranks and selection budgets}\label{bidirectional-ranks-and-selection-budgets}

Each protein query ranked the same 711 molecules, and each molecule query ranked the same 614 proteins. For each of the 78 unordered system pairs, we calculated Spearman correlation from average ranks within each query, then averaged over queries. Cross-system summaries report the median of these system-pair means. Candidate scores were non-constant within every query, so all within-query Spearman correlations were defined. Comparisons share systems and queries, so the system pairs are descriptive units rather than independent replicates.

We evaluated \(K=5\), 10 and 20. Candidates above the Top-\(K\) boundary received membership weight 1 and those below it received 0. If a tied group of size \(b\) crossed the boundary after \(a\) higher-scoring candidates, each tied candidate received weight \((K-a)/b\). The inner product of two membership vectors is the expected intersection under independent uniform tie breaking. For two independent uniform selections from \(N\) candidates, expected overlap is \(K^2/N\). The probability that two untied Top-10 subsets are disjoint is \(\binom{N-10}{10}/\binom{N}{10}\). Observed disjointness requires a zero weighted intersection, so the lists are disjoint under every admissible tie resolution. The resulting proportion is therefore a lower bound on the probability of disjoint lists under random tie breaking. These combinatorial references describe recommendation overlap.

Candidate concentration was the fraction of Top-10 slots occupied by a system's ten most frequently recommended candidates.

\subsection{Exact query reassignment and conditional references}\label{exact-query-reassignment-and-conditional-references}

Let \(a_q\) and \(b_q\) denote centred unit-norm rank vectors, or Top-\(K\) membership vectors, for query \(q\). Uniformly permuting the query identities of one system while retaining candidate identities gives the exact reference

\[
R=\mathbb{E}_\pi\left[\frac{1}{Q}\sum_q a_q^T b_{\pi\left(q\right)}\right]=\overline{a}^T\overline{b}.
\]

The difference between observed agreement and \(R\) is the query-aligned agreement increment \(\Delta\),

\[
\Delta=Q^{-1}\sum_q\left(a_q-\overline{a}\right)^{\top}\left(b_q-\overline{b}\right).
\]

The reference averages over all query permutations, including fixed points. It preserves persistent candidate preferences and the ordering within each query. The increment quantifies query-aligned excess agreement between systems. Different query-dependent responses can also produce a small increment.

Conditional references permuted queries within sequence-similarity components, broad experimental classes or exact Bemis--Murcko scaffold groups. We retained groups with at least two queries and used the same queries for observed and reassigned values. MMseqs2 sequence groups required at least 30\% identity, at least 80\% coverage on both sequences and an E-value of at most 0.001 \citep{mmseqs2}. Groups were the connected components of the resulting similarity graph. Molecular scaffolds ignored chirality, and acyclic molecules formed individual singleton groups \citep{rdkit_scaffold}.

\subsection{Persistent preferences and sensitivity analyses}\label{persistent-preferences-and-sensitivity-analyses}

For a centred, unit-normalised query rank vector \(z_q\),

\[
z_q=\overline{z}+\left(z_q-\overline{z}\right).
\]

The mean vector represents persistent candidate preference, and the residual represents query-dependent change. Query-dependent rank variation was \(V=1-\|\overline{z}\|^2\). Dividing a system pair's query-aligned agreement increment by the geometric mean of its two \(V\) values gives the cosine similarity between their query-dependent rank residuals. This separates the amount of query-dependent variation from its alignment across systems.

In a five-fold grouped procedure, we fitted a static candidate ranking from mean percentile ranks in four folds and evaluated Top-10 recovery on the fifth. Protein-homology or molecular-scaffold groups remained within folds. The query-blind ranking used model scores only.

We assessed sampling variation with 2,000 resamples of query groups drawn with equal probability and replacement. We normalised each resampled mean by the number of queries in the selected groups and kept the candidate pool fixed. To assess candidate-pool sensitivity, we used 20 fixed seeds per direction to form nested pools of 128, 256 and 512 candidates and the full pool of 711 molecules or 614 proteins. Rankings and tie weights were recomputed within each score subset. Chance-adjusted overlap was \((O_K-K^2/N)/(K-K^2/N)\). MAMMAL pKd and BALM were also evaluated on a separate 128 × 128 catalogue with no shared molecular or protein identities with the primary catalogue (Supplementary S3).

\subsection{Cumulative candidate coverage}\label{cumulative-candidate-coverage}

For each number of systems \(m\), we averaged the union of Top-\(K\) recommendations over queries and all equally weighted \(m\)-system subsets. Boundary ties contributed membership probabilities, with tie breaking independent across systems. For the reassignment reference, we independently permuted whole query lists within each system, including fixed points. Its exact expectation used each system's mean membership probabilities across queries. Uniform expected coverage was \(N\left[1-\left(1-K/N\right)^m\right]\). We evaluated \(K=5\), 10 and 20 in the thirteen-system set and the nine-system sensitivity subset (Supplementary S3).

\subsection{Controlled training and recommendation reproducibility}\label{controlled-training-and-recommendation-reproducibility}

The controlled model used curated BindingDB and ChEMBL data with a fixed molecular-scaffold split comprising 337,570 training, 41,806 validation and 39,507 test pairs. Training included 105,368 explicitly inactive pairs and 192,469 observations with exact dissociation constants (Kd) or inhibition constants (Ki) for auxiliary regression. A shared interaction network used frozen 768-dimensional molecular features from BerMol \citep{lu2025dtiam} and 1,280-dimensional ESM2 protein features, with each projected to 192 dimensions. Pairs were ranked by the mean of two classification logits. All four objectives retained balanced binary cross-entropy and Huber regression for the Kd and Ki endpoints with weight 0.1. The objectives differed only in ranking supervision from measured positive and negative pairs: no ranking loss, molecule-selection ranking with weight 0.2, protein-selection ranking with weight 0.2, or both directions with weight 0.1 each. Unknown pairs were not treated as negatives.

Each objective used seeds 101, 202 and 303, giving twelve fits. Within each seed, objectives shared initialisation, per-step training inputs and forward-pass order, including auxiliary inputs with zero loss weight. The primary comparison used exponential-moving-average model weights at step 5,000. For a secondary comparison, we selected the model state with the highest validation mean of the two directional query-macro AP values. These values averaged AP across eligible queries in the quantitative Kd/Ki observations. Evaluation included AP over all test pairs, query-macro AP for quantitative Kd/Ki queries containing both positive and negative observations, and false-positive rates on explicit inactives using validation-selected thresholds. Top-10 agreement was measured on a fixed 468-molecule × 469-protein screening catalogue with features available from both encoders. Supplementary S8 gives the network, loss functions, optimisation and stopping rules, query eligibility and all seed results.

A further factorial experiment comprised 27 fits with fixed classification and regression labels. It varied the measured pairs eligible for ranking supervision, screening direction and query weighting while matching the number of forward-pass inputs per training step. We examined the functional consequences of seed variation on the 70 × 65 overlap panel. Pairs used for training or validation-based model selection were identified separately for the controlled model (Supplementary S8--S9).

\subsection{Retrospective functional panel}\label{retrospective-functional-panel}

Functional measurements came from the published kinase inhibitor profiling data associated with KIRHub \citep{kirhub2026}. The source reports residual kinase activity at 1 µM. We calculated percentage inhibition as 100 minus percentage residual activity and matched 5,372 measurements for 79 molecules and 68 targets to source Data Table S4. Source Data Table S2 identified two complexes, CDC7--DBF4 and CDK7--cyclin H--MNAT1, and a non-default MAPK8 isoform. Excluding these three columns gave the complete 79 × 65 primary panel. Sensitivities retained all 68 targets or restricted the primary panel to 39 targets described as full length. Target associations used gene and accession identifiers, and compound associations used the assay names and catalogue structures. Supplementary S1 documents molecular identity checks. Source construct and expression descriptions and affinity tags were retained as assay annotations.

The primary hit criterion was at least 70\% inhibition at 1 µM, with 50\% and 80\% sensitivities. The main budget was five candidates per query. We also assessed budgets of 1, 3 and 10. All queries were retained, including those with no measured hits. Each selection rule used the same candidate pool. Fractional membership at tied boundaries kept the total selection weight equal to the per-query budget. The measured endpoint is inhibition at 1 µM. Source summary data did not include replicate-level error estimates.

A source-overlap sensitivity excluded the union of exact pair matches to the two audited BindingDB source pools from every system. Using only the resulting eligibility pattern, we retained all 65 kinases and the 57 molecules with measurements eligible for every kinase. The same frozen scores, hit threshold, budget and query-group assignments were used on this complete subpanel (Supplementary S1 and S4).

\subsection{Candidate-selection rules and hit-yield decomposition}\label{candidate-selection-rules-and-hit-yield-decomposition}

We formed five outer query-group folds from molecular-scaffold or protein-homology groups without using activity values. Within each development set, four-fold grouped validation selected one of thirteen systems or an adjustment strength \(\alpha\in\{0,0.25,0.5,1\}\) by mean hit yield at budget five and the 70\% inhibition threshold. Ties favoured the smaller \(\alpha\) or the alphabetically first system. Evaluation queries were excluded from preference fitting and selection-rule fitting. Candidate identities were shared across query folds.

Let \(t_m(q,c)\) be the within-query percentile rank. Scores were ranked in ascending order with average ties, and rank \(r\) was transformed as \((r-1)/(C-1)\) for \(C\) candidates. Selection rules were a query-blind ranking from development-query mean preference, mean percentile rank, summed Top-10 membership votes followed by selection to budget, preference-adjusted combination \(C_\alpha(q,c)=\operatorname{mean}_m\left[t_m(q,c)-\alpha\mu_m(c)\right]\), and the development-selected single system. Candidate preference \(\mu_m(c)\) used development queries only. The threshold and budget sensitivity analyses retained the selected rules.

For a final recommendation vector \(w_q\) and measured hit vector \(y_q\), matched hit yield was \(G=\operatorname{mean}_q\left(w_q^T y_q\right)\). We reassigned whole recommendation lists within each outer evaluation fold, preserving dependencies within the final list. The exact fold reference was \(\operatorname{mean}\left(w\right)^{T}\operatorname{mean}\left(y\right)\), averaged across folds by query count and including fixed points. The outcome reference uses evaluation measurements. The query-blind rule uses only model scores from development queries. Uniform hit yield \(U\) was the budget multiplied by the measured positive fraction. Thus,

\[
G=U+(R-U)+(G-R).
\]

The two components are enrichment retained after reassignment, \(R-U\), and correspondence gain, \(G-R\). Both \(G\) and \(G-R\) were reported because changes in the latter can arise from either matched or reassigned yield. Conditional percentile intervals used 2,000 paired query-group resamples of per-query contributions, with fitted selection rules, references and candidate identities fixed. They quantify variation across query groups without refitting models or resampling assay measurements.

\subsection{Strict crossed selectivity}\label{strict-crossed-selectivity-1}

We enumerated all crossed-selectivity units with at least 70\% inhibition on both preferred pairs and at most 20\% on both crossed pairs. A stricter sensitivity used at least 80\% and at most 10\%. Protein-selection recovery required both molecules to rank their preferred protein higher. Molecule-selection recovery required both proteins to rank their preferred molecule higher. Strict recovery required all four inequalities, with ties counted as failures and also reported separately.

We evaluated the thirteen fixed systems on all 26,114 qualifying units. Comparisons of fitted selection rules used 832 units whose molecules shared an outer fold and whose proteins shared an outer fold, so each direction used one fixed rule within a unit. Direction-specific fitted rules need not form a single scalar score matrix. Fixed combinations were additionally summarised over all units. Because units share entities, we report entity counts and leave-one-scaffold- or sequence-group-out ranges (Supplementary S4).

For four independent, identically distributed continuous scalar scores, the 24 strict orders are equiprobable. Strict recovery places the two preferred-pair scores above the two crossed-pair scores, giving \(2!\times2!/4!=1/6\). This mathematical reference applies to a common scalar score matrix, rather than separately fitted directional rules. An additive score \(s(d,p)=a(d)+b(p)\) or its strictly increasing transform cannot reverse either candidate preference across queries, so its strict recovery rate is zero.

\subsection{Representation and selection diagnostics}\label{representation-and-selection-diagnostics}

Supplementary S5 evaluates activity ranking and low-entropy selection on the functional panel. Auxiliary inhibition-regression and hit-probability models use molecular-scaffold and protein-homology group holdouts, with standardisation and fitting confined to each training block. Fig.~\ref{fig:5} and Supplementary S6--S7 analyse model-specific PCA, common-anchor representations, linear readouts fitted to predict model scores and synthetic scaling of the doubly centred pair residual; for dual-tower systems without joint pair embeddings (ConPLex, DrugCLIP and BALM), anchor similarities were evaluated using product kernels over separate protein and ligand representations. Calibration and evaluation identities were separated on both axes for the geometry and readout comparisons. PCA, anchor scaling and linear readouts were determined from calibration pairs. Analyses used scores paired with the extracted representations. The selected representation layers, kernels, regularisation and uncertainty calculations are specified with each analysis.

\subsection{Study scope}\label{study-scope}

The released-system analysis compares specified inference configurations on shared candidates. The functional analysis compares the retrospective recommendation behaviour of deployments with different training histories. These comparisons do not separate the effects of training exposure from other deployment differences or establish relative generalisation to unseen pairs. The kinase panel measures retrospective hit yield and strict crossed selectivity. Controlled training examines variation within one learning setup. Source-data overlap and configuration sensitivities are reported in Supplementary S1, with outcome, representation and training sensitivities in S4--S9.

\section{Author contributions}\label{author-contributions}

Jiangjie Qiu led the study and contributed throughout, including study conception and design, methodology, software, analysis, visualisation and manuscript preparation. Chengrui Xu performed representation-related analyses, organised computational experiments, and contributed to figure preparation and manuscript writing. Xiaonan Wang supervised the study, provided resources and scientific guidance, and contributed to refining the overall research design.

\section{Conflicts of interest}\label{conflicts-of-interest}

The authors declare no competing interests.

\section{Data availability}\label{data-availability}

Model predictions, processed evaluation data and source data for the figures and tables are supplied with the manuscript for peer review. The accompanying code reproduces the primary ranking, system-level hit-yield and strict crossed-selectivity analyses, and baseline seed comparisons from the fixed input data. The data and custom code will be deposited in a persistent repository upon acceptance and made publicly available with a DOI upon publication. Original datasets and released model implementations are identified in Methods and the Supplementary Information.

\section{Acknowledgements}\label{acknowledgements}

This work was supported by the Shenzhen Science and Technology Program (AI2026004), the Tsinghua University Initiative Scientific Research Program and the Scientific Research Innovation Capability Support Project for Young Faculty (ZYGXQNJSKYCXNLZCXM-E7). OpenAI Codex was used to assist with programming and manuscript preparation.
\clearpage
\rsccontrol
\bibliographystyle{rsc}
\bibliography{references}
\end{document}